\documentclass{article}
\usepackage{ijcai21}

\usepackage{times}

\usepackage{soul}
\usepackage{url}
\usepackage{enumitem}
\usepackage[hidelinks]{hyperref}
\usepackage[utf8]{inputenc}
\usepackage[small]{caption}
\usepackage{graphicx}
\usepackage{amsmath}
\usepackage{amsfonts,amssymb}
\usepackage{booktabs}
\DeclareMathOperator*{\argmin}{arg\,min}
\newtheorem{assumption}{Assumption}
\newtheorem{definition}{Definition}
\newcommand{\indep}{\rotatebox[origin=c]{90}{$\models$}}
\title{Causal Learning for Socially Responsible AI}

\author{
Lu Cheng\footnote{Contact Author}\and
Ahmadreza Mosallanezhad\thanks{Equal contribution}\and
Paras Sheth\footnotemark[2]\And
Huan Liu\\
\affiliations
Computer Science and Engineering, Arizona State University
\emails
\{lcheng35, amosalla, psheth5, huanliu\}@asu.edu
}

\begin{document}

\maketitle

\begin{abstract}
There have been increasing concerns about Artificial Intelligence (AI) due to its unfathomable potential power. To make AI address ethical challenges and shun undesirable outcomes, researchers proposed to develop socially responsible AI (SRAI). One of these approaches is causal learning (CL). We survey state-of-the-art methods of CL for SRAI. We begin by examining the seven CL tools to enhance the social responsibility of
AI,  then review how existing works have succeeded using these tools to tackle issues in developing SRAI such as fairness. The goal of this survey is to bring forefront the potentials and promises of CL for SRAI. 
\end{abstract}
\section{Introduction}
Artificial Intelligence (AI) comes with both promises and perils. AI significantly improves countless aspects of day-to-day life by performing human-like tasks with high efficiency and precision. It also brings potential risks for oppression and calamity because how AI works has not been fully understood and regulations surround its use are still lacking \cite{cheng2021socially}. Many striking stories in media (e.g., Stanford's COVID-19 vaccine distribution algorithm) have brought \textit{Socially Responsible AI} (SRAI) into the spotlight.  

Substantial risks can arise when AI systems are trained to improve accuracy without knowing the underlying \textit{data generating process} (DGP). First, the societal patterns hidden in the data are inevitably injected into AI algorithms. The resulting socially indifferent behaviors of AI can be further exacerbated by data heterogeneity and sparsity. Second, lacking knowledge of the DGP can cause researchers and practitioners to unconsciously make some frame of reference commitment to the formalization of AI algorithms \cite{getoor2019responsible,cheng2021socially}, spanning from data and label formalization to the formalization of evaluation metrics. Third, DGP is also central to identifying the cause-effect connections and the causal relations between variables, which are two indispensable ingredients to achieve SRAI. We gain in-depth understanding of AI by intervening, interrogating, altering its environment, and finally answering ``what-if'' questions. 

Causal inference is the key to uncovering the real-world DGPs \cite{pearl2009causality}. In the era of big data, especially, it is possible to learn causality by leveraging both causal knowledge and the copious real-world data, i.e., \textit{causal learning} (CL) \cite{guo2020survey}. There have been growing interests seeking to improve AI's social responsibility from a CL perspective, e.g., causal interpretability \cite{moraffah2020causal} and causal-based machine learning fairness \cite{makhlouf2020survey}. In this survey, therefore, we first examine the seven tools in CL that are inherently related to SRAI. We then review existing efforts of connecting four of these tools to emerging tasks in SRAI, including \textit{bias mitigation, fairness, transparency}, and \textit{generalizability/invariance}. We conclude with open problems and challenges of leveraging CL to enhance both the functionality and social responsibility of AI. 
\begin{figure*}
    \centering
    \includegraphics[width=.85\linewidth]{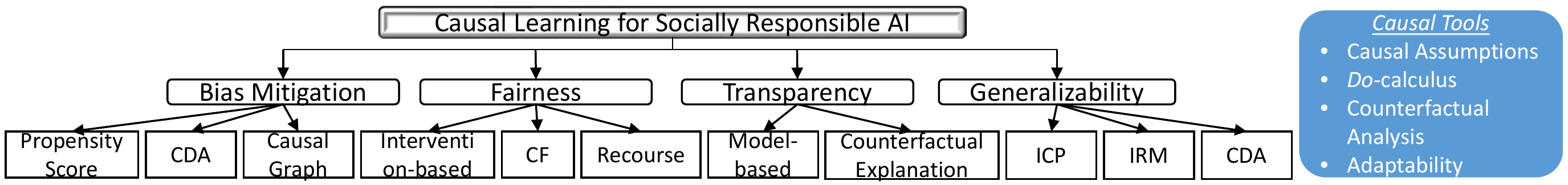}
    \caption{The taxonomy of CL for SRAI. The blue rectangle denotes the four commonly used causal tools for SRAI.}
    \label{fig:taxonomy}
\end{figure*}
\section{Causal Tools for Socially Responsible AI}
Based on the available causal information, we can describe CL in a three-layer hierarchy: association, intervention, and counterfactual \cite{pearl2019seven}. At the first layer, association seeks statistical relations between variables, i.e., $p(y|x)$. By contrast, intervention and counterfactual demand causal information. An intervention is a change to the DGP. With $do$-calculus \cite{pearl2009causality}, the interventional distribution $p(y|do(x))$ describes the distribution of $Y$ if we force $X$ to take the value $x$ while keeping the rest in the process same. This corresponds to removing all the inbound arrows to $X$ in the causal graph. At the top layer is counterfactual, denoted as $p(y_x|x',y')$. It stands for the probability of $Y=y$ had $X$ been $x$ given what we observed were $X=x'$ and $Y=y'$. 

In the rest of this section, we review the seven tools of CL introduced in \cite{pearl2019seven} and briefly discuss how it naturally steers a course towards a SRAI-future.
\begin{itemize}[leftmargin=*]
    \item \textit{Causal Assumptions} make an AI system more transparent and testable. Encoding causal assumptions explicitly allows us to discern whether these assumptions are plausible and compatible with available data. It also improves our understandings of how the system works and gives a precise framework to debate \cite{cloudera2020causality}.
    \item \textit{Do-calculus} enables the system to exclude spurious correlations by eliminating confounding through back-door criterion \cite{pearl2009causality}. Confounding is a major cause of many socially indifferent behaviors of an AI system.
    \item \textit{Counterfactual analysis} involves a ``what would have happened if'' question. It is the building block of scientific thinking, thus, a key ingredient to warrant SRAI. 
    \item \textit{Mediation Analysis} decomposes the effect of an intervention into direct and indirect effects. Its identification helps understand how and why a cause-effect arises. Therefore, mediation analysis is essential for generating explanations.
    \item \textit{Adaptability} is a model's capability to generalize to different environments. AI algorithms typically cannot perform well when the environment changes. CL can uniquely identify the underlying mechanism responsible for the changes. 
    \item \textit{Missing data} is a common problem in AI tasks. It reduces statistical power, data representativeness, and can cause biased estimations. CL can help recover causal and statistical relationships from incomplete data via causal graphs. 
    \item \textit{Causal discovery} learns causal directionality between variables from observational data. As causal graphs are mostly unknown and arbitrarily complex in real-world applications, causal discovery offers a ladder to true causal graphs. 
\end{itemize}
We propose the taxonomy of CL for SRAI in Fig. \ref{fig:taxonomy} and review how these CL tools -- particularly, the Causal Assumptions, Do-calculus, Counterfactual Analysis, and Adaptability -- help SRAI in terms of bias mitigation, fairness, transparency, and generalizability/invariance below.
\section{Bias Mitigation}
\label{s:bias_mitigation}
AI systems can be biased due to hidden or neglected biases in the data, algorithms, and user interaction. \cite{olteanu2019social} introduced 23 different types of bias including selection, measurement biases, and so on. There are various ways to de-bias AI systems such as adversarial training~\cite{zhang2018mitigating}, reinforcement learning~\cite{wang2020mitigating}, and causal inference~\cite{zhao2019gender}. Due to its interpretable nature, causal inference offers high confidence in making decisions and can show the relation between data attributes and AI system's outcomes. Here, we review two popular causality-based methods for bias mitigation -- propensity score and counterfactual data augmentation (CDA).
\subsection{Propensity Score}
Propensity score is used to eliminate treatment selection bias and ensure the treatment and control groups are comparable. It is the ``conditional probability of assignment to a particular treatment given a vector of observed covariates'' \cite{rosenbaum1983central}. Due to its effectiveness and simplicity, propensity score has been used to reduce unobserved biases in various domains, e.g., NLP and recommender systems. Here, we focus our discussions on recommender systems. 

Inverse Propensity Scoring (IPS) is used to alleviate the selection and position biases commonly present in recommender systems. Selection bias appears when users selectively rate or click items, rendering observed ratings not representative of the true ratings, i.e., ratings obtained when users randomly rate items. Given a user-item pair $(u,i)$ and $O_{u,i}\{0,1\}$ denoting whether $u$ observed $i$, we define propensity score as $P_{u,i}=P(O_{u,i}=1)$, i.e., the marginal probability of observing a rating. During the model training phase, IPS-based unbiased estimator is defined using following empirical risk function \cite{schnabel2016recommendations}:
\begin{equation}
\small
    \argmin_{\theta} \sum_{O_{u,i}=1}\frac{\hat{\sigma}_{u,i}(r,\hat{r}(\theta))}{P_{u,i}}+Reg(\theta),
\end{equation}
where $\hat{\sigma}_{u,i}(r,\hat{r}(\theta))$ denotes an evaluation function and $Reg(\theta)$ the regularization for model complexity. IPS is also used to mitigate selection bias during evaluation, see, e.g., \cite{schnabel2016recommendations,yang2018unbiased}. 

Position bias occurs in a ranking system as users tend to interact with items with higher ranking positions. To remedy position bias, previous methods used IPS to weigh each data instance with a position-aware value. The loss function of such models is defined as follows~\cite{agarwal2019general}: $L(M, q) = \sum_{x\in \pi_q} \Delta(x, y | \pi_q),$
where $M$ indicates the ranking model, $q \in Q$ denotes a query from a set of all queries to the model, $\pi_q$ is the ranked list by $M$, and $\Delta(x,y|\pi_q)$ denotes the individual loss on each item $x$ with relevant label $y$. In another method, \cite{hofmann2013reusing} proposed to estimate propensity scores using ranking randomization. First, the ranking results of the system are randomized. Then the propensity score is calculated based on user clicks on different positions. Although this method is shown to be effective, it can significantly degrade user experience as the highly ranked items may not be users' favorites~\cite{joachims2017unbiased}. Therefore, \cite{guo2020debiasing} proposed a method in an offline setting where randomized experiments are not available. They specifically considered multiple types of user feedback and applied IPS to learn an unbiased ranking model.
\subsection{Counterfactual Data Augmentation (CDA)} 
CDA is a technique to augment training data with their counterfactually-revised counterparts via causal interventions that seek to eliminate spurious correlations \cite{kaushik2020learning}. It enables AI algorithms to be trained on unseen data, therefore, reducing undesired biases and improving model generalizability. Here, we focus on CDA for bias mitigation and will discuss model generalizability in Sec. 6.3.

One of the domains using CDA to reduce biases hidden in the data is NLP. Particularly, one begins by sampling a subset of original data that contains attributes of interest, e.g., gender or sentiment. Then expert annotators or inference models are asked to generate the counterfactual counterparts. The augmented datasets are later fed into the downstream NLP tasks, e.g., sentiment analysis. For example, CDA can be used to reduce gender bias by generating a dataset that encourages training algorithms not to capture \textit{gender}-related information. One such method generates a gender-free list of sentences using a series of sentence templates to replace every occurrence of gendered word pairs (e.g., he:she, her:him/his) \cite{lu2020gender}. It formally defined CDA as:
\begin{definition}
Given input instances $S = \{(x_1, y_1), (x_2, y_2), ..., $ $ (x_N, y_N)\}$ and intervention $c$, a $c$-augmented dataset $S'$ is $S \cup \{(c(x), y)\}_{(x,y)\in S}$.
\end{definition}
The underlying assumption is that an unbiased model should not distinguish between matched pairs and should produce the same outcome for different genders~\cite{lu2020gender}. For sentiment analysis, \cite{kaushik2020learning} generated counterfactual scenarios with help of human annotators who were provided with positive movie reviews and were asked to perform minimal changes to make them negative.
\begin{figure}
    \centering
    \includegraphics[width=.5 \linewidth]{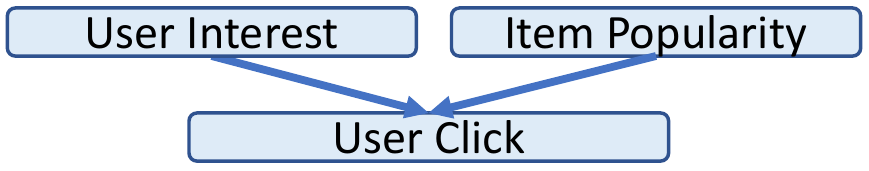}
    \caption{A causal graph to mitigate popularity bias.}
    \label{fig:causal_graph_zheng}
\end{figure}
\subsection{Causal Graphs}
Causal graph is a powerful approach for counterfactual reasoning as it allows us to study the effects of various sensitive attributes on the outcome. It is a probabilistic graphical model to encode assumptions and widely used to detect and mitigate bias in AI systems. In the NLP domain, for example, to mitigate gender bias in word embeddings, a recent work by \cite{yang2020causal} proposed a causal graph characterizing relation between gender-definition and gender-biased non-gender-definition word vectors. In the domain of recommender systems, causal graph can mitigate bias that affects decisions toward popular items (i.e., popularity bias) or gender-specific recommendations. The underlying assumption is that, a user click on biased recommendation results from two independent causes: user interest and item popularity. Or formally, $P_{\text{click}} = P_{\text{user interest}} + P_{\text{item popularity}},$
where $P$ indicates matching probability for a user and an item. A corresponding causal graph that disentangles user interest and item popularity is shown in Fig. \ref{fig:causal_graph_zheng}. They further created two different embeddings to capture users' real-interest in items and pseudo-interest caused by popularity. Finally, the user interest embedding is used to create a de-biased recommender where the popularity bias has been disentangled.

\subsection{Discussions}
Bias is a primary reason that AI systems fail to make fair decisions. Using propensity-score-based approaches needs to specify the correct forms of propensity scores in real-world applications. The alternate randomization experiments -- which may be inapplicable due to ethical and financial considerations -- might decrease the utility performance of the AI systems~\cite{joachims2017unbiased}. One challenge of using CDA is to design a process to generate a modified dataset using the considered interventions. While causal graphs appear promising~\cite{yang2020causal}, one has to make assumptions that may be impractical to reflect real-world biases. Beyond bias mitigation, measuring bias via experimentation can help understand causal connections between attributes of interest and algorithmic performance \cite{balakrishnan2020towards}, therefore, mitigating biases. 
\section{Fairness}
Biases in AI systems can lead to many undesired consequences such as model overfitting and other societal issues. One of the most frequently discussed issues in AI is fairness, the property of a model that produces results independent of given variables, especially those considered sensitive, e.g., gender. Here, we briefly review another line of research that aims to train a fair AI system apart from the de-biasing perspective discussed above. A comprehensive survey on causal fairness can be referred to \cite{makhlouf2020survey}.
\subsection{Fairness via Causal Modeling}
From the causal perspective, fairness can be formulated as estimating causal effects of sensitive attributes such as gender on the outcome of an AI system. Such causal effects are evaluated using counterfactual interventions over a causal graph with features, sensitive attributes, and other variables. Underpinning this approach is the concept of \textit{counterfactual fairness} (CF)~\cite{kusner2017counterfactual}. CF implies that a decision is considered fair if it is the same in both ``the actual world'' and ``a counterfactual world'' where, e.g., for an individual belongs to a different demographic group. Formally, considering $Y$, $A$, and $X$ as the observed outcome, sensitive attributes, and features, respectively, CF is defined as follows:
\begin{definition}
Given $x\in X$ and $a \in A$, predictor $\hat{Y}$ is counterfactually fair if
\begin{equation}
\small
     \begin{aligned}
        &P(\hat{Y}_{A\leftarrow a}(U)=y|X=x, A=a) \\&=  P(\hat{Y}_{A\leftarrow a'}(U)=y|X=x, A=a)
    \end{aligned}
\end{equation}
   holds for all $y$ and any $a' \in A$.
\end{definition}
$U$ refers to a set of latent background variables in a causal graph. This definition states that if the two outcome probabilities $P(y_{a}|x, a)$ and $P(y_{a'}|x, a)$ are equal for an individual, then s/he is treated fairly as if s/he had been from another sensitive group. Or, $A$ should not be the cause of $\hat{Y}$. CF assumed that fairness can be uniquely quantified from observational data, which is not valid in certain situations due to the unidentifiability of the counterfactual quantity. \cite{wu2019counterfactual} then introduced a graphical criterion determining the identifiability of counterfactual quantities. With the derived bounds, CF is achieved by constraining the classifier's training process on the amount of unfairness in the predictor.

Another stepping stone toward creating fair AI systems using CL is the intervention-based fairness. \cite{hajian2012methodology} first proposed to improve fairness by expressing direct and indirect discrimination through the different paths connecting the sensitive attributes (i.e., the treatment) and the outcome of an AI system. Similarly, \cite{zhang2018fairness} modeled discrimination based on the effect of a sensitive attribute on an outcome along certain disallowed causal paths. These methods generally formulate and quantify fairness as the average causal effect of the sensitive attribute on the decision attribute, which is then evaluated by the intervention through the post-interventional distributions.
\subsection{Discussions}
CF considers more general situations than intervention-based fairness where the set of profile attributes is empty, therefore, it is more challenging. One limitation of CF is that these sensitive attributes may not admit counterfactual manipulation~\cite{kasirzadeh2021use}. What does it mean to suppose a different version of an individual with the counterfactually manipulated sensitive attributes such as race? In order to use a sensitive attribute appropriately, it is necessary to specify what the categories are and what perception of an attribute to be used. The validity of counterfactuals is also challenging to assess. What is the metric used to measure the similarity between the actual and imaginary worlds? CF-based models may fail when the identifiable assumption is violated due to the unidentifiable counterfactual quantity~\cite{wu2019counterfactual}. Although this might be solved via a relaxed assumption, the idea of using social categories to achieve fairness can be problematic as they may not admit counterfactual manipulation. Lastly, parallel to algorithmic fairness, algorithmic \textit{recourse} offers explanations and recommendations to individuals unfavourably treated. Future research can consider causal algorithmic recourse and its relation to CF and other fairness criteria \cite{von2020fairness}.
\section{Transparency}
Conventional AI algorithms often lack \textit{transparency}, typically presented in the concept of \textit{interpretability/explanability}. When AI algorithms do not provide explanations for how and why they make a decision, users' trust on these algorithms can be eroded. 
Hence, there is a need for AI systems to produce interpretable results. Causal Interpretability helps generate human friendly explanations by answering questions such as ``Why does a model makes such decisions?''. In this section, we describe two approaches associated with causal interventional interpretability and counterfactual interpretability, respectively. Please refer to \cite{moraffah2020causal} for more details.
\subsection{Model-based Interpretations}
CL for model-based interpretations seeks to estimate the causal effect of a particular input neuron on a certain output neuron in the network. Drawing from causal inference theories, causally interpretable models first map a neural network structure into a Structural Causal Model (SCM) \cite{pearl2009causality} and then estimate the effect of each model component on the output based on the data and a learned function (i.e., a neural network) using $do$-calculus \cite{chattopadhyay2019neural}. Particularly, every $l$-layer neural network $N(l_1,l_2,...,l_n)$ has a corresponding SCM $M([l_1,...,l_n],U,[f_1,...,f_n],P_U)$ where $f_i$ refers to the set of causal functions for neurons in layer $l_i$. $U$ denotes a group of exogenous random variables that act as causal factors for input layer $l_1$. $P_u$ defines the probability distribution of $U$. $M$ can be further reduced to a SCM with only input layer $l_1$ and output layer $l_n$: $M'([l_1,l_n],U,f',P_U)$, by marginalizing out the hidden neurons \cite{chattopadhyay2019neural}. Finally, we can estimate the average causal effect (ACE) of a feature $x_{i} \in l_1$ with value $\alpha$ on output $y\in l_n$ by 
\begin{equation}
ACE_{do\left(x_{i}=\alpha\right)}^{y}=\mathbb{E}\left[y \mid do\left(x_{i}=\alpha\right)\right]-baseline_{x_{i}},
\end{equation}
where $baseline_{x_{i}} = \mathbb{E}_{x_{i}}\left[\mathbb { E } _ { y } \left[y \mid do\left(x_{i}=\right.\right.\right. \alpha)]]$.

Similar method was applied to CNN architectures trained on image data to reason over deep learning models \cite{narendra2018explaining}. \cite{zhao2021causal} leveraged partial dependence plot \cite{friedman2001greedy} and Individual Conditional Expectation \cite{goldstein2015peeking} to extract causal information (e.g., relations between input and output variables) from black-box models. \cite{martinez2019explaining} proposed an approach for explaining the predictions of a visual model with the causal relations between the latent factors which they leveraged to build a Counterfactual Image Generator.
\subsection{Counterfactual Explanations} 
Different from model-based interpretation which deals with model parameters to determine the vital components of the model, counterfactual explanations typically describe scenarios such as ``If X had not occurred, Y would not have occurred''. Specifically, the predicted outcome is considered as the event $Y$ and the features fed to the model are the causes $X$. A counterfactual explanation can be defined as a causal situation of the form where an output $Y$, which occurs given the feature input $X$, can be changed to a predefined output $Y'$ by minimally changing the feature vector $X$ to $X'$.

To generate counterfactual explanations, a common approach is a generative counterfactual framework \cite{liu2019generative} that leverages generative models along with attribute editing mechanisms. The objective function is defined as
\begin{equation}
\label{Eq1}
\small
\begin{array}{l}
\underset{x_{c f}}{\arg \min } \max _{\lambda} L\left(x, x_{c f}, y, y_{c f}\right) \\
L\left(x, x_{c f}, y, y_{c f}\right)=\lambda \cdot\left(\hat{f}\left(x_{c f}\right)-y_{c f}\right)^{2}+d\left(x, x_{c f}\right),
\end{array}
\end{equation}
where $x$/$x_{cf}$ denotes the observed/counterfactual features and $y$/$y_{cf}$ the observed/counterfactual outcome. The first term indicates the distance between the model’s prediction for the counterfactual input $x_{cf}$ and the desired counterfactual output. The second term describes the distance between the observed features $x$ and counterfactual features $x_{cf}$. $\lambda$ is the hyperparameter balancing the importance of the two distances. Another category of approach relies on adversarial examples to provide counterfactual explanations. Rather than explaining why a model predicts an output with a given input, it finds an alternative version of the input that receives different classification results. One can also lay constraints on the features so that only the desired features are subject to changes. The third category of approach uses class prototypes for the counterfactual search process \cite{van2019interpretable}. These prototypes -- refer to the mean encoding of the instances that belong to the class -- are integrated into the objective function so that the perturbations can produce interpretable counterfactuals. For a more detailed understanding readers could refer to \cite{xu2020causality}.

\subsection{Discussions}
There are still limitations in existing models for causal interepretability. For instance, evaluating such models is difficult due to the lack of ground-truth causal relations between the components of the model or causal effect of one component on another. While using counterfactual explanations for interpretability may seem feasible, it has been shown that there exist unspecified contextual presumptions and choices while generating counterfactual scenarios that may not stand well in the social world \cite{kasirzadeh2021use}. For instance, social categories such as race and gender need to be first defined in order to generate counterfactual scenarios around these variables. Even with ground truth and well-defined social categories, existing works may still fail because causal assumptions in these works are not explicitly explained. It is critical to clearly define causal assumptions. 
\section{Invariance/Generalizability}
Due to societal biases hidden in data and the shortcut learning \cite{geirhos2020shortcut}, AI algorithms can easily overfit to training data, i.e., learning \textit{spurious correlations}. Common approaches for avoiding overfitting rely on the assumption that samples of the entire population are i.i.d., which is rarely satisfied in practice. Violating the i.i.d. assumption leads to poor generalizability of an AI algorithm. Because whether a training and a testing DGP (or environment) differ is unknown, we have recourse to data heterogeneity and a model that is robust to distributional shifts among heterogeneous environments, or the \textit{invariance} property \cite{arjovsky2019invariant}. Causal relations are, by their nature, invariant \cite{pearl2011transportability}. Environment is defined by intervention, therefore, in a causal graph, only direct causal relations remain invariant when an intervention changes the environment. We first examine two popular approaches that incorporate the invariance property into predictions. 
\subsection{Invariant Causal Prediction (ICP)}
Built upon SCM, ICP \cite{peters2016causal} aims to discover the causal parents (direct causes) of a given variable directly pointing to the target variable without constructing the entire causal graph. We consider the setting where multiple environments $e\in \mathcal{E}$ exist and in each environment $e$, there is a predictor variable $X^e\in\mathbb{R}^p$ and a target variable $Y^e\in\mathbb{R}$. Given a set $S\subseteq\{1,...,p\}$, a vector $X_{S}$ containing all variables $X_k, k\in S$, ICP assumes Invariant Prediction:
\begin{assumption}[Invariant Prediction.] There is a vector of coefficients $\gamma^*=(\gamma_1^*,...,\gamma_p^*)^\intercal$ with support $S^*:=\{k:\gamma_k^*\neq0\}\subseteq\{1,...,p\}$ such that for all $e\in\mathcal{E}$ and $X^e$ with an arbitrary distribution:
\begin{equation}
\small
    Y^e=\mu+X^e\gamma^*+\epsilon^e, \quad \epsilon^e\sim F_{\epsilon}, \epsilon^e\indep X_{S^*}^e,
\end{equation}
where $\mu$ is the intercept and $\epsilon^e$ denotes the random noise with the same distribution $F_{\epsilon}$ across all $e\in \mathcal{E}$.
\end{assumption}

\noindent With multiple environments, ICP then fits a linear (Gaussian) regression in each environment. The goal is to find a set of features that results in invariant predictions between environments. In particular, ICP iterates over subsets of features combinatorially and looks for features in a model that are invariant across environments, i.e., invariant coefficients or residuals. The intersection of these sets of features is then a subset of the true direct causes. ICP also relies on the unconfoundedness assumption \cite{pearl2009causality}: no unobserved confounders exist between input features and the target. In practice, it is common to choose an observed variable to be the environment variable (e.g., background color of an image), when it could plausibly be so. Limited to using linear models and discrete variable for environment separation, \cite{heinze2018invariant} extended ICP to a non-linear setting.  
\subsection{Invariant Risk Minimization (IRM)}
Causal graphs are, in many cases, inaccessible, e.g., the causal relations between pixels and a target predict. Without the need to retrieve direct causes of a target variable in a causal graph, IRM \cite{arjovsky2019invariant} elevates invariance by focusing on \textit{out-of-distribution} (OOD) generalization -- the performance of a predictive model when evaluated in a new environment. IRM seeks to learn a data representation $\phi$ that achieves two goals: predicting accurately and eliciting an invariant predictor across environments. This can be formulated as the constrained optimization problem:
\begin{equation}
\small
\min_{\phi:\mathcal{X}\rightarrow\mathcal{Y}}\sum_{e\in\mathcal{E}}R^e(\phi)+\lambda\cdot\|\bigtriangledown_{w|w=1.0}R^e(w\cdot\phi)\|^2,
\end{equation}
where $\phi$ is the invariant predictor in a latent causal system generating observed features, $w=1.0$ is a fixed ``dummy'' classifier. $R^e(\cdot)$ denotes the risk under $e$ such as prediction errors. $\lambda$ controls the balance between prediction accuracy in each $e$ and the invariance of the predictor $1\cdot\phi(x)$. In practice, the prediction performance in the defined environments is almost certainly reduced due to the exclusion of some spurious correlations. IRM cannot guarantee to remove all spurious correlations as it also depends on the provided environments. There are a number of follow-up works such as \cite{jin2020domain,krueger2020out} relying on a stronger assumption of invariance of $p(y|\phi(x))$ than that of $\mathbb{E}[y|\phi(x)]$ in IRM.
\begin{figure}
    \centering
    \includegraphics[width=\linewidth]{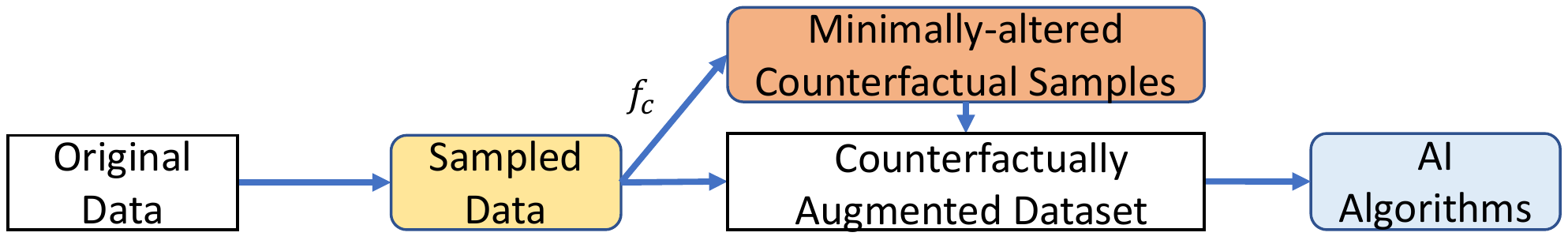}
    \setlength{\belowcaptionskip}{-10pt}
    \caption{CDA for Invariance. $f_c$ is a counterfactual distribution.}
    \label{fig:my_label}
\end{figure}
\subsection{CDA for Invariance/Generalizability}
Another causality-inspired approach for improving model invariance is to augment original data with counterfactuals that can expose the model to OOD scenarios. CDA prevents models from learning spurious patterns present in the training data, thus, improving model invariance. A general pipeline describing CDA for model invariance can be seen in Fig. \ref{fig:my_label}.

In computer vision, CDA is used to generate counterfactual images close to training samples yet may not belong to existing training categories. For example, to detect unknown classes, generative adversarial networks were used to generate perturbed examples from the known class and labeled as unknown categories \cite{neal2018open}. Drawing on independent mechanisms (IMs) \cite{peters2017elements}, \cite{sauer2021counterfactual} proposed to decompose the image generation process into different mechanisms related to its shape, texture, and background. With known causal structure and learned IMs, a counterfactual generative network generated counterfactual images regarding each mechanism. Similar concept was used in visual question answering \cite{abbasnejad2020counterfactual} to improve the generalizability of various multimodal and unimodal vision and language tasks. 

Inherently related to causal inference, Reinforcement Learning (RL) uses CDA to learn more generalizable policies. Counterfactual data in RL introduces various scenarios that an RL agent generally does not experience during training. In dynamic processes, for instance, \cite{pitis2020counterfactual} proposed to decompose the dynamics of different subprocesses into local IMs which can be used to generate counterfactual experiences. To choose the optimal treatment for a given patient, \cite{lu2020sample} proposed a data-efficient RL algorithm that used SCM to generate counterfactual-based data. 
\subsection{Discussions}
Current applications of IRM have been focused on computer vision, nevertheless, an environment needs not to be scenery in an image. Some promising applications include health care \cite{kouw2018introduction}, robotics \cite{giusti2015machine}, NLP \cite{choe2020empirical}, recommender systems \cite{wang2018deconfounded}, and so on. IRM also highly relates to fairness \cite{arjovsky2019invariant}. When applying IRM, one may pay attention to the non-linear settings where formal results for latent-variable models are lacking and risks are under-explored \cite{rosenfeld2020risks}. Another caveat of existing works in invariant prediction is the reliance on the stringent unconfoundedness assumption, which is typically impractical. ICP is more interpretable than IRM in terms of discovering causal features. For CDA, the counterfactual samples generation strategy usually relieves the conditional independence assumption of training data, which helps improve model generalizability. When generating counterfactual data is not feasible, one can use minimally-different examples in existing datasets with different labels to improve model generalizability \cite{teney2020learning}.

\section{Summary and Open Problems}
We review recent advances in SRAI from CL perspective. Purely reliant on statistical relationships, current AI algorithms achieve prominent performance meanwhile its potential risks raise great concerns. To achieve SRAI, we argue that CL is an effective means for it seeks to uncover the DGPs. Our survey begins by introducing the seven CL tools and their connections to SRAI. We then discuss how four of these tools are used in developing SRAI. In the following, we briefly describe promising future research directions of SRAI.
\noindent\paragraph{Privacy-preserving.} Privacy is a crucial tenet of SRAI. Many research has shown that AI systems can learn and remember users' private attributes. However, how to use CL to enhance privacy has been barely studied in literature. Similar to de-biasing methods, we can use CL to remove sensitive information and create privacy-preserving data representations.
\noindent\paragraph{Making explicit causal assumptions.} Explicitly making assumptions ensures more valid, testable, and transparent causal models. Given causal assumptions might be disputed or uncertain, we need sensitivity analysis to measure the model performance with assumption violations. Critically, assumptions should be made with humility and researchers are responsible to protect against unethical assumptions.
\noindent\paragraph{Causal discovery.} While causal discovery has been extensively studied, its connection to SRAI is not well understood. Discovering causal relations helps determine if assumptions are properly made and interventions are correctly applied. Given that causal graph is key to many CL approaches in SRAI, causal discovery is an important future research. 
\noindent\paragraph{Mediation analysis.} Causal mediation analysis improves model transparency. For example, in CF, sensitive attributes such as gender and race are assumed to solely have direct influence on the classification. Is the effect of race on loan granting mediated by the job type? Similarly, mediation analysis could be used in explainable AI, e.g., neurons directly or indirectly influence algorithmic decisions.
\noindent\paragraph{Missing data.} CL is a missing data problem: inferring the potential outcomes of the same units with different treatment assignments. We might apply CL to a more general setting of missing data. For example, graphical model based procedures can be used to provide performance guarantees when data are Missing Not At Random \cite{mohan2021graphical}. 
\noindent\paragraph{Long-term impact.} The majority of works in SRAI overlooks its long-term commitment to be fulfilled. This hinders both the efficiency and efficacy of existing works to achieve SRAI. For instance, static fairness criterion used in bank loan granting may cost credibility scores of the minorities in the long run \cite{liu2018delayed}. 
\noindent\paragraph{Social good.} Essentially, SRAI is designed to \textit{protect, inform} users, and \textit{prevent/mitigate} the harms of AI \cite{cheng2021socially}. With the burgeoning AI-for-social-good movement, CL is becoming the core component of AI systems to tackle societal issues. 
\noindent\paragraph{Causal tools and libraries for SRAI.} SRAI research can also benefit from using existing CL libraries such as Causal ML\footnote{https://github.com/uber/causalml}, DoWhy\footnote{https://microsoft.github.io/dowhy}, and Causal Discovery Toolbox\footnote{https://fentechsolutions.github.io/CausalDiscoveryToolbox/html}. It is possible to integrate CL models for SRAI into these tools.
\section*{Acknowledgements}
This material is based upon work supported by, or in part by, the U.S. Army Research Laboratory and the U.S. Army Research Office under contract/grant number W911NF2110030 and W911NF2020124 as well as by the National Science Foundation (NSF) grant 1909555.
\bibliographystyle{named}  
\bibliography{ijcai21}
\end{document}